\documentclass[conference]{IEEEtran}
\IEEEoverridecommandlockouts
\usepackage[bookmarks=false]{hyperref}
\usepackage{cite}
\usepackage{amsmath,amssymb,amsfonts}
\usepackage{graphicx}
\usepackage{textcomp}
\usepackage{xcolor}
\usepackage{mathtools}
\usepackage{amsmath}
\usepackage{amssymb}
\usepackage{bbm}
\usepackage{bm}
\usepackage{bbold}
\usepackage{algorithm,algcompatible,amsmath}

\usepackage{subcaption}

\begin{document}

\title{Discovering Subdimensional Motifs of Different Lengths in Large-Scale Multivariate Time Series}

\author{
    \IEEEauthorblockN{Yifeng Gao\IEEEauthorrefmark{1}, Jessica Lin\IEEEauthorrefmark{1}}
    
    \IEEEauthorblockA{\IEEEauthorrefmark{1}Department of Computer Science, 
     George Mason University, Virginia, USA
     \\
     \{ygao12, jessica\}@gmu.edu}
\vspace{-10mm}
}

\maketitle

\begin{abstract}
Detecting repeating patterns of different lengths in time series, also called variable-length motifs, has received a great amount of attention by researchers and practitioners. Despite the significant progress that has been made in recent single dimensional variable-length motif discovery work, detecting variable-length \textit{subdimensional motifs}---patterns that are simultaneously occurring only in a subset of dimensions in multivariate time series---remains a difficult task. The main challenge is scalability. On the one hand, the brute-force enumeration solution, which searches for motifs of all possible lengths, is very time consuming even in single dimensional time series. On the other hand, previous work show that index-based fixed-length approximate motif discovery algorithms such as random projection are not suitable for detecting variable-length motifs due to memory requirement. In this paper, we introduce an approximate variable-length subdimensional motif discovery algorithm called \textbf{C}ollaborative \textbf{HI}erarchy based \textbf{M}otif \textbf{E}numeration (CHIME) to efficiently detect variable-length subdimensional motifs given a minimum motif length in large-scale multivariate time series. We show that the memory cost of the approach is significantly smaller than that of random projection. Moreover, the speed of the proposed algorithm is significantly faster than that of the state-of-the-art algorithms. We demonstrate that CHIME can efficiently detect meaningful variable-length subdimensional motifs in large real world multivariate time series datasets.

\end{abstract}

\begin{IEEEkeywords}
Variable Length, Motif Discovery, Time Series
\end{IEEEkeywords}

\section{Introduction}

Detecting repeating patterns of various lengths, also called variable-length motifs, in time series has received a great amount of attention \cite{li2012visualizing}\cite{mueen2013enumeration}\cite{hime}\cite{gao2017trajviz}. Since motifs of different lengths can naturally co-exist in a time series, detecting variable-length motifs often is a necessary step for many real-world applications such as classification \cite{mueen2013enumeration}, anomaly detection\cite{wang2016self} and data visualization \cite{balasubramanian2016discovering}.



Contrary to the significant progress that has been made in recent single dimensional variable-length motif discovery work \cite{mueen2013enumeration}\cite{hime}\cite{Gao2018}, only little progress is made in detecting variable-length \textit{subdimensional motifs} \cite{minnen2007detecting}\cite{yeh2017matrix} --- patterns that are simultaneously occurring only in a subset of all dimensions in multivariate time series. 

Existing approaches \cite{minnen2006discovering}\cite{minnen2007detecting}\cite{yeh2017matrix} in subdimensional motif discovery still only detect motifs of a specified length, possibly suggested by domain experts. While in some applications, these approaches may fit well if domain knowledge is available and a good motif length can be specified by the user, we aim at solving the problem in a more general case --- when the correct motif length is not known, or motifs of various lengths co-exist in the data.

We illustrate the limitation of fixed-length motif discovery in Figure 1. In the figure, there are two subdimensional motifs of lengths 200 and 400 respectively. The first subdimensional motif (labeled in red with green box) occurs in the first two dimensions $D_1$ and $D_2$, and the second subdimensional motif occurs in $D_2$ and $D_3$. Since the motifs have different lengths, even in the best case, existing (fixed-length) subdimensional motif discovery approaches \cite{minnen2006discovering}\cite{minnen2007detecting}\cite{yeh2017matrix} can only detect one of these motifs correctly if the proper length is provided. When the lengths are unknown, they would need to try different lengths --- a process that is very time consuming. To overcome this limitation, our proposed approach is designed to discover motifs of various lengths in a single run; that is, \textit{all} subdimensinoal motifs can be discovered even with significant length differences between them.

\begin{figure}[t]
 \centering
 \includegraphics[width=75mm]{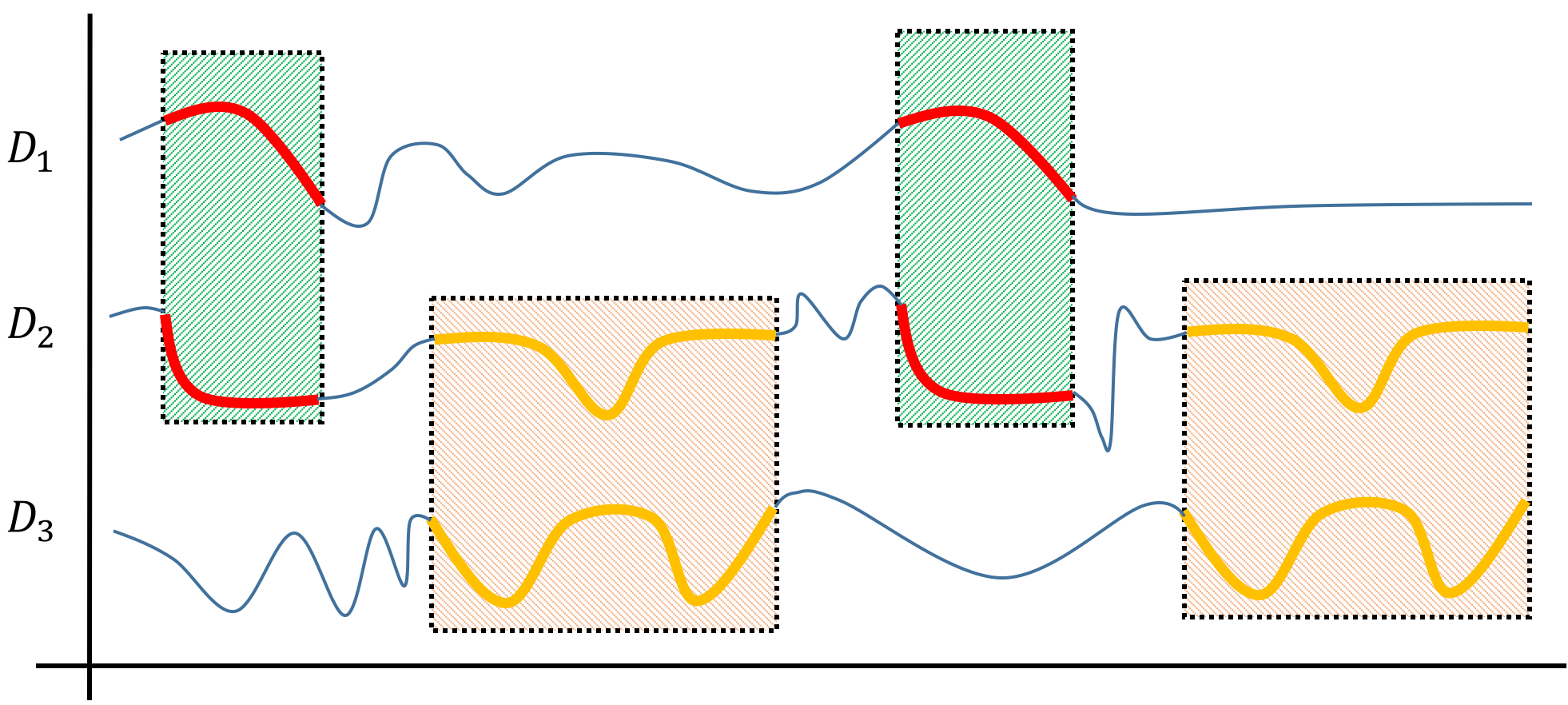}
 \tiny\caption{Example of two subdimension motif. One is labeled in red line and occurs in dimension $\{D_1,D_2\}$ with length 200. The other one appears in $\{D_2,D_3\}$ with length 400.}
 \label{fig:example}
 \vspace{-4mm}
\end{figure}

Subdimensional motifs of different lengths widely exist in real world time series data. For example, the Caltrans Performance Measurement (PEMS) dataset \cite{pems}, which records traffic information in major California cities over time, can have patterns in different spatial and temporal scales ranging from several hours of rush hour traffic patterns to a 5-day stable weekly pattern. In meteorology data analysis, a motif with duration of several hours may benefit short-term forecasting, whereas a motif that lasts several months may represent some seasonal phenomenon \cite{yeh2017matrix}. These motifs can co-exist in the same time series even though they differ significantly in length, and span different dimensions. Detecting them is often necessary for analyzing the data \cite{mueen2013enumeration}\cite{valmod}.

The main challenge for detecting variable-length subdimensional motif is scalability. The brute-force enumeration solution, which searches for motifs of all possible lengths, is very time consuming. While there exist some search techniques for detecting single dimensional variable-length motifs \cite{mueen2013enumeration}\cite{valmod}\cite{itr}, they cannot be generalized to detecting subdimensional motifs since they do not address the problem of searching relevant dimensions. As a result, how to efficiently detect variable-length subdimensional motifs remains a difficult problem.


In this paper, we introduce an algorithm called \textbf{C}ollaborative \textbf{HI}erarchy based \textbf{M}otif \textbf{E}numeration (CHIME) to efficiently detect variable-length subdimensional motifs given a minimum motif length $l$. Towards that end, CHIME combines repeating symbols detected from each dimension, and uses the symbol matching results in all dimensions collaboratively to avoid redundant searches. While the algorithm does not guarantee an exact solution, it can find motifs with considerably large length range in multivariate time series with high accuracy. Even in a fifty-dimensional time series with length of one million, the algorithm executes in less than one hour. This is significantly faster than the state-of-the-art approaches which would take days to detect motifs of a \textit{single} fixed length. The efficient search makes variable-length subdimensional motif discovery in large-scale multivariate time series a feasible task. To summarize, our work has the following contributions:

\begin{itemize}
    \item The algorithm can discover subdimensional motifs of a large length range. 
 
    \item The speed is significantly faster than that of the state-of-the-art algorithms. 
    
     \item Even though CHIME is an approximate algorithm, the experiments show that the algorithm can detect motifs with high accuracy.
    
    \item The experiments also show that CHIME can discover meaningful motifs in real world datasets, and can potentially benefit other multivariate time series data mining tasks such as classification.
\end{itemize}

The rest of the paper is organized as follows: Sec. 2 discusses related work and challenges in detecting variable-length subdimensional motifs. Sec. 3 introduces the problem definition and notations used in the paper. Sec. 4 describes the discretization technique and introduces our algorithm. The experimental results are shown in Sec. 5, and we conclude in Sec. 6.

\section{Related Work}

We start by introducing recent \textit{fixed-length} multivariate time series motif discovery work.

Tanaka et al. \cite{tanaka2005discovery} introduce a smart approach that compresses multivariate time series into a univariate time series to detect motifs at a low cost. Minnen et al. and Berlin et al.  \cite{minnen2006discovering}\cite{berlin2012detecting} utilize a density based approach to locate potential motif areas. All three approaches above, however, are designed for finding patterns that match in all dimensions. Therefore, the performance of these approaches will degrade if there exist some irrelevant dimensions.

Another work by Minnen et al. \cite{minnen2007detecting} is perhaps the first work to point out that finding patterns that match in all dimensions is not very useful in some real-world applications. The authors show that explainable motifs often occur in only some subset(s) of all dimensions. They introduce a fast approach that utilizes random projection and Symbolic Aggregate approXimation representation (SAX) to efficiently find approximate subdimensional motifs. Following similar idea, Wang et al. \cite{wang2010tree} introduce a subdimensional motif discovery approach by constructing a suffix tree. Recently, Yeh et al. \cite{yeh2017matrix} introduce an algorithm named mSTAMP that utilizes state-of-the-art motif discovery results along with Minimum Description Length (MDL) metric to locate subdimensional motifs. The approach can detect high-quality, meaningful subdimensional motifs in many real-world datasets. 

All the algorithms described above require the user to pre-define the motif length. However, recent work have shown that detecting motifs of different, possibly unknown lengths is necessary in many real-world applications \cite{hime}\cite{itr}\cite{mueen2013enumeration}\cite{valmod}. We next discuss recent \textit{variable-length} subdimensional motif discovery algorithm.

Presently, work on variable-length subdimensional motif discovery is very limited. We only found one such work by Balasubramaian et al. \cite{balasubramanian2016discovering}. The authors utilize a Grammar Induction based framework \cite{li2012visualizing} to find variable-length subdimensional motifs in health-care time series. However, their approach requires storing every combination of the co-occurring single dimensional motifs in memory. So while it can achieve high accuracy, the approach is only suitable for low-dimensional time series. In contrast, our proposed algorithm is able to scale in high dimensional, large-size time series.

\section{Notation \& Definition}

We start with fundamental definitions related to time series:

\textbf{Single Dimensional (Univariate) Time Series} $T=t_1,\dots,t_N$ is a set of observations ordered by time. 

\textbf{Multivariate Time Series} $\mathbf{T}=T_1,T_2,\dots,T_D$ is a set of $D$ co-evolving single dimensional time series $T_i$. 

\textbf{Subsequence} $s_{j,p,q}=[t_{j,p},\dots,t_{j,q}]$ of a multivariate time series $\mathbf{T}$ is a contiguous set of points in the univariate time series $T_j$ starting from position $p$ with length $L=q-p+1$. Typically, $L << N$ and $1\leq p \leq N-L+1$. 

Subsequences can be extracted from $T_i$ via a sliding window. In many applications, we are interested in finding similar ``shapes.'' Therefore, motif discovery is more meaningful when it is offset- and amplitude-invariant. This can be achieved by normalizing each subsequence prior to the search for motifs. \textbf{Z-normalization} is a procedure that normalizes the mean and standard deviation of all points in the subsequence to zero and one, respectively. 

Given a start and an end position $p$, $q$ respectively, a \textbf{Multidimensional Subsequence} consisting of $D$ subsequences can be extracted. We denote it as $\mathbf{S}_{\mathbf{D},p,q}$ where $\textbf{D}$ contains all possible dimensions $\{1,2,\dots,D\}$.

As demonstrated in previous work \cite{minnen2007detecting}\cite{yeh2017matrix}, in many cases only a subset of all co-evolving subsequences in $\mathbf{S}_{\mathbf{D},p,q}$ are relevant (i.e. have repeating patterns). So we describe the concept of subdimensional subsequence \cite{yeh2017matrix} used in previous work.

The \textbf{Subdimensional Subsequence} $\mathbf{S}_{\mathbf{d},p,q}$ is a set of subsequences among $D$ subsequences that start from position $p$ and end at $q$ in multivariate time series $\mathbf{T}$, where $\mathbf{d}$ is an indicator vector that stores the list of relevant dimensions and $|\mathbf{d}|$ is the number of relevant dimensions. 




We next introduce the definition of subdimensional motif used in this work. Given a distance function $Dist(.,.)$ and two subdimensional subsequences $\mathbf{S}_{\mathbf{d},p1,q1}$ and  $\mathbf{S}_{\mathbf{d},p2,q2}$ with the same indicator $\mathbf{d}$ and the same length, a vector $dist_{i}$ is used to record the distance value $Dist(s_{j,p1,q1},s_{j,p2,q2})$ where $j \in \mathbf{d}$. In this work, we define a \textbf{Subdimensional Motif} to be a set of subdimensional subsequences in $\mathbf{T}$ such that the average value of the distance vector $dist$  between the subdimensional subsequences and a seed subdimensional subsequence $\mathbf{S}_{seed}$ is less than the motif threshold function $R(L)$, where $L$ is the length of the motif detected, $L \geq l$ (the minimum motif length). Each such subdimensional subsequence is said to be an instance of the subdimensional motif.

\subsection{Problem Definition}

Finally, we introduce the problem that the proposed approach aims to address.

\textbf{Variable-Length Subdimensional Time Series Motif Discovery Problem}: Given a minimum motif length $l$ and a multivariate time series $\mathbf{T}$, the problem aims to find \textit{all} variable-length subdimensional time series motifs that exist in $\mathbf{T}$. 

Clearly, finding exact solution for this problem in a large-scale multivariate time series is costly. However, discovering even a subset of variable-length motifs can already benefit many real-world applications compared with fixed-length motifs \cite{mueen2013enumeration}\cite{li2012visualizing}\cite{Gao2018}. Therefore, given a minimum motif length $l$, our proposed algorithm aims to find an approximate set of high quality variable-length motifs with lengths larger than $l$, and rank them based on some interest measure.

It is worth noting that determining a general interest measure for variable-length subdimensional motif discovery problem is non-trivial \cite{yeh2017matrix}\cite{minnen2007detecting}\cite{valmod}. Some long motifs are likely to have larger distances compared with short motifs even after normalizing the distances by length. Some motifs that have a lower number of dimensions may contain more interesting local patterns compared with those spanning most or all of the dimensions. Therefore, the ranking of subdimensional motifs should be based on the application. The reader may refer to \cite{yeh2017matrix}\cite{minnen2007detecting}\cite{valmod} for details on various motif evaluation approaches for different goals. 

In our experiments, since most state-of-the-art approaches evaluate motifs based on similarity, when compared with those techniques, we rank the motifs based on the most similar pair of subsequences within each detected motif, and we include the execution time of pairwise comparison \cite{mueen2009exact}\cite{mueen2013enumeration}\cite{yeh2017matrix}. 

\section{Proposed Method}



In this section, we first describe Symbolic Aggregate approXimation (SAX), the discretization approach we use to represent each subsequence. Then we describe the Multivariate Time Series Numerosity Reduction process, the pre-processing step that aims to remove neighboring, similar subsequences to avoid over-counting a pattern. We then introduce the proposed method. 

\subsection{Symbolic aggregate approximation (SAX)}

Discretization \cite{lin2007experiencing} is a common step in many time series motif discovery algorithms \cite{lin2007experiencing}\cite{hime}\cite{chiu2003probabilistic}\cite{minnen2007detecting} \cite{balasubramanian2016discovering}\cite{li2012visualizing}, for it often helps improve the efficiency of the algorithms significantly. In this section, we describe Symbolic Aggregate approXimation (SAX), a popular technique used to discretize univariate time series. 

Given a normalized subsequence from $\mathbf{T}$ and a reduced dimension size $w$, SAX converts the raw subsequence into lower-dimenional PAA (Piecewise Aggregate Approximation) representation by dividing the subsequence into $w$ equal-sized windows, and computing the average value of the points within each window. Intuitively, the PAA coefficients vector \cite{lin2007experiencing} is a $w$-dimensional vector that consists of 
the average values from $w$ equal-sized segments of the input subsequence. PAA coefficients can be considered as an approximate representation of the original subsequence. 

Given a PAA coefficients vector, the algorithm then maps it to $w$ symbols with alphabet size $a$ according to a breakpoint table \cite{lin2007experiencing}, defined such that the regions are approximately equal-probable under Gaussian distribution. This maximizes the chance that the symbols occur with approximately equal probability. These $w$ symbols form a SAX word. Figure 2 summarizes the process. The bold flat lines represent the values of PAA coefficients computed from their respective segments in the subsequence; the breakpoint table with $a$ from 2 to 4 is also shown in the figure. Since we set $a=3$, the second column shows the breakpoints, based on which three regions are generated: $[-\infty,-0.43),[-0.43,0.43),[0.43,\infty]$. The PAA coefficients falling into these three regions are represented by symbols $a,b$ and $c$ respectively. In this example, the SAX word $abca$ is formed.


\begin{figure}[h]
 \centering
 \includegraphics[width=70mm]{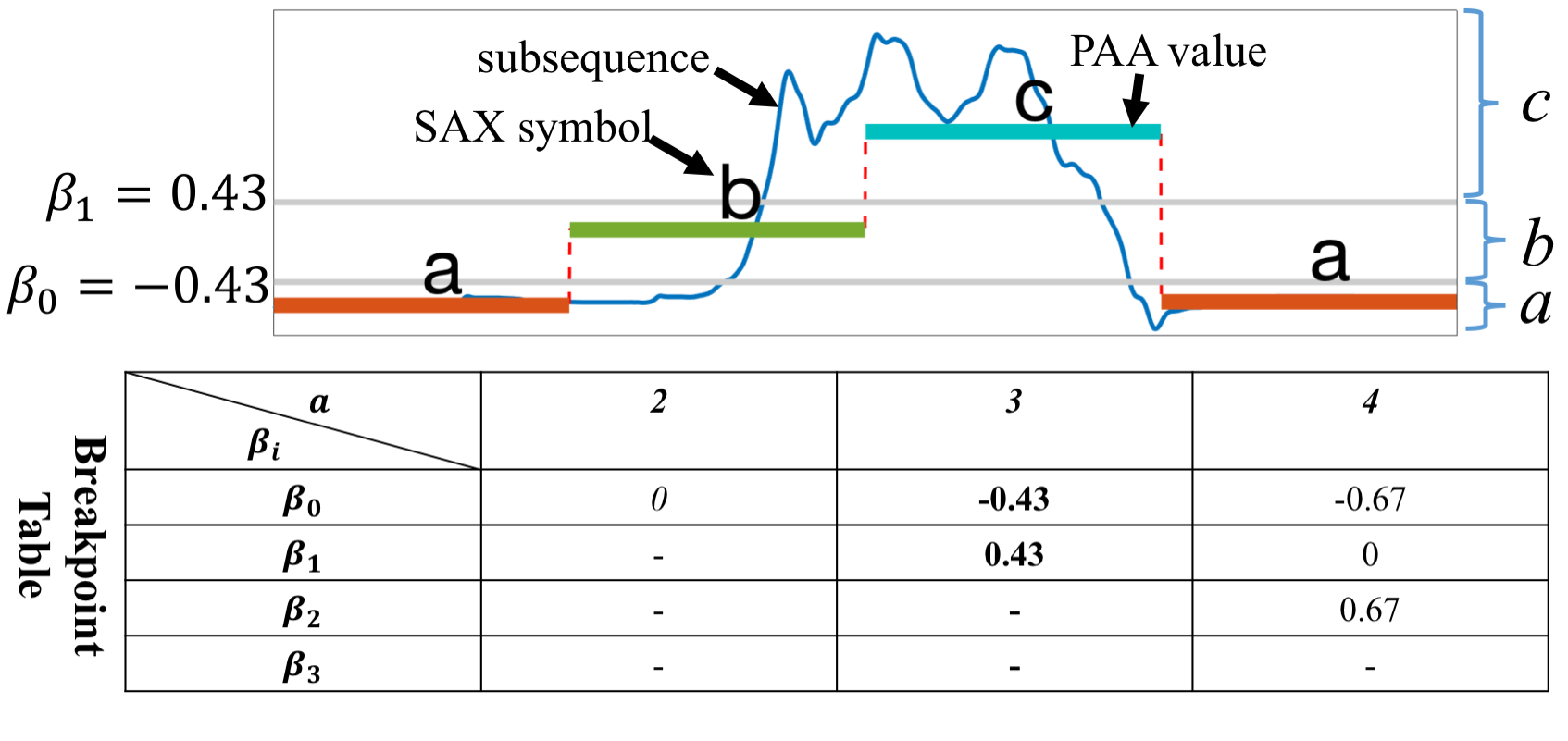}
 \caption{Example of Generating SAX word ``abca'' with $a=3$ and $w=4$. }
 \label{fig:sax:example}
\end{figure}

\subsubsection{Fast Computation of SAX for Subsequence of Any Length}

Since the proposed approach uses SAX words heavily, it is very important to reduce the time cost of discretization. We next describe a fast algorithm to compute the SAX word for a univariate subsequence of any length \cite{hime}. We first pre-compute two vectors of statistical features for every time series $T_i \in \mathbf{T}$: $ESum_x(x,d)=\sum_{i=1}^{x}t_{i,d}$ and $ESum_{xx}(x,d)=\sum_{i=1}^{x}t_{i,d}^2$ based on input multivariate time series $\mathbf{T}$. Given a subsequence $s_{p,q,d}$ of length $q-p+1$ in dimension $d$, the SAX representation can be computed by Algorithm 1. In the algorithm, the mean and variance of $s_{p,q,d}$ is computed in constant time (Line 3-4). The computation cost of computing PAA coefficients (Line 5-7) is $O(w)$, which is not dependent on subsequence length. The PAA coefficients are converted to SAX word based on the pre-defined breakpoint table described in previous subsection, which takes $O(wa)$ time computation. The overall cost of the whole process is $O(w+wa)$. Since $Esum_x(x,d),Esum_{xx}(x,d)$ can be computed during pre-possessing in $O(ND)$ time, the cost of computing a SAX word for arbitrary length subsequence during motif discovery process is independent of the subsequence length. As demonstrated in \cite{lin2007experiencing}\cite{hime}, $w$ and $a$ should be very small compared with subsequence length. So the time cost to compute a SAX word is significantly reduced ($w+wa\ll L$).

\begin{algorithm}[h]
    \caption{Fast SAX Computation (FastSAX)}
  \begin{algorithmic}[1]
    \STATE \textbf{Input}: $Esum_x$,$Esum_{xx}$, PAA size $w$, subsequence $s_{p,q,d}$
    \STATE \textbf{Output}: PAA representation $paa$
    \STATE $Ex=Esum_x(q,d)-Esum_x(p,d)$ 
    \STATE $E_{xx}=Esum_{xx}(q,d)-Esum_{xx}(p,d)$
    \STATE $L=q-p+1$,  $\mu=\frac{E_{x}}{L}$, $\sigma=\sqrt{\frac{E_{xx}-E_{x}^2/L}{L-1}}$
    \FOR{every PAA segment}
     \STATE $paa_i=(\frac{Esum_{x}(paa_{i,e},d)-Esum_{x}(paa_{i,s}),d}{L/w}-\mu)/\sigma$
    
     \Comment{$paa_{i,s}$ and  $paa_{i,e}$ is the start and end point of each PAA segments}
     \ENDFOR
     \STATE \textbf{return} ConvertPAAToSAX($paa$)
  \end{algorithmic}
\end{algorithm}

\subsection{PAA Distance-based Multivariate Time Series Numerosity Reduction}

In practice, neighboring subsequnces extracted via a sliding window are similar to each other since they are off by one point. To avoid over-counting a pattern, and to allow variable-length pattern discovery, Numerosity Reduction often is used to further compress the word sequence in previous univariate variable-length motif discovery work \cite{li2012visualizing}\cite{hime}\cite{Gao2018}. 


Concretely, given a multivariate time series $\mathbf{T}$, similar to previous work \cite{li2012visualizing}\cite{hime}\cite{Gao2018}, PAA Distance-based Numerosity Reduction process conducts a left to right scan through $\mathbf{T}$ and only records the first multivariate subsequence $s_{\mathbf{D},p,q}$ such that the PAA distance\footnote{Given two subsequence, PAA distance is computed by the distance between the PAA coefficient vectors of two subsequences multiplied by a factor of $\sqrt{l/w}$. The PAA distance has been shown to lower bound the actual Euclidean Distance. \cite{lin2007experiencing}. The reader can refer to the previous work \cite{lin2007experiencing} for detailed explanation on PAA distance.}
\cite{lin2007experiencing} between $s_{\mathbf{D},p,q}$ and the most recently recorded subsequence \cite{lin2007experiencing} is greater than $2R(l)$ in at least one dimension. Similar to previous work \cite{Gao2018}\cite{hime}, we set $w=32$ for computing the PAA distance.

\subsection{Collaborative Hierarchy based Motif Enumeration (CHIME)}

In this section, we describe our proposed method. 

\subsubsection{Basic Data Structure}

\begin{figure}[h]
\vspace{-2mm}
 \centering
 \includegraphics[width=80mm]{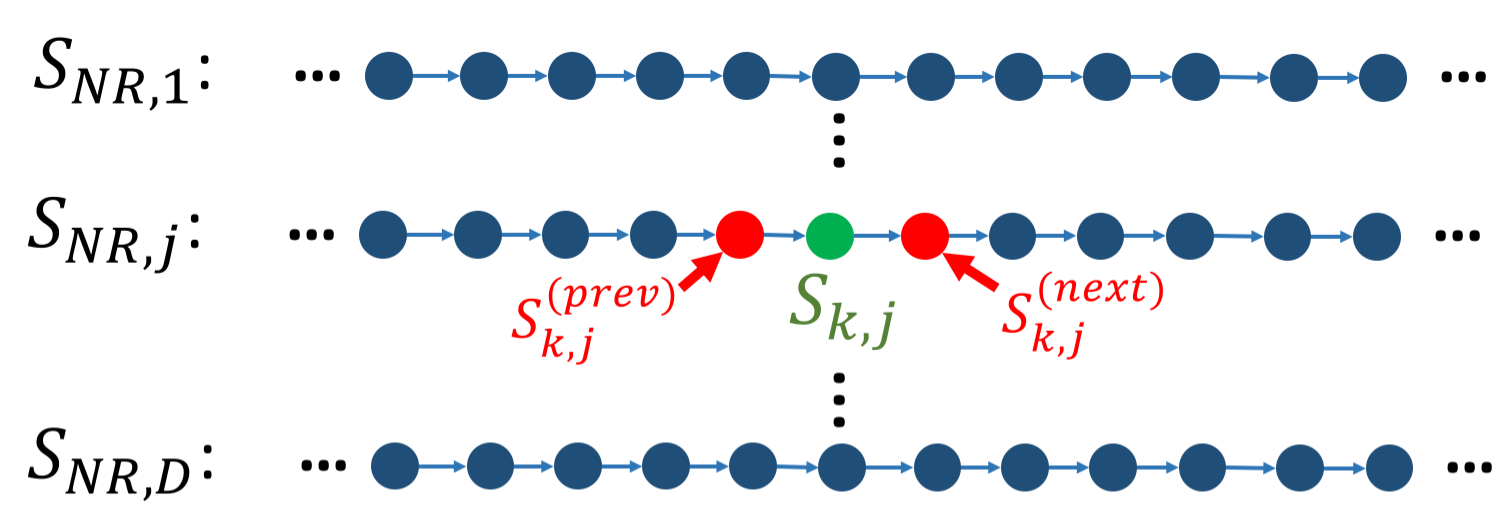}
 \caption{Data Structure used for proposed approach.}
 \label{fig:ind}
\vspace{-3mm}
\end{figure}

Since the proposed work heavily relies on recursively building up hierarchical structure based on the numerosity reduced subsequences, we store all subsequences after Numerosity Reduction in $D$ linked lists (denoted as $\mathcal{S}_{NR}$). Specifically, for each single dimensional time series $T_j \in \mathbf{T}$, the reduced subsequences are stored in a linked list $\mathcal{S}_{NR,j}$ shown in Fig. \ref{fig:ind}.

Each node $S_{k,j} \in \mathcal{S}_{NR,j}$ stores a subsequence $s_{j,p,q}$ and is connected to two nodes which represent the two subsequences that appear before and after $s_{j,p,q}$ in the reduced subsequence sequence, respectively. We use $S_{k,j}^{(next)}$ and $S_{k,j}^{(prev)}$ to represent these two nodes respectively. The edge that connect $S_{k,j}$ with $S_{k,j}^{(next)}$ is denoted as the \textit{next} edge. The algorithm will check every subsequence connected by the \textit{next} edge, and merge the connected nodes to generate longer motifs. 

Once the data structure is built, CHIME conducts a left to right scan for every node in $\mathcal{S}_{NR,j} \in \mathcal{S}_{NR}$ and calls Collaborative Enumerator --- the algorithm used to enumerate variable-length subdimensional motifs based on node $S_{k,j} \in \mathcal{S}_{NR,j}$.

\subsubsection{Collaborative Enumerator}



Collaborative Enumerator is described in Alg 2. Intuitively, given a node $S_{k,j}$ (Line 1-2), the enumerator recursively grows the length of the subsequence by using a greedy enumeration step and a dimension matching step. These two steps collaboratively avoid redundant enumeration, which saves space cost, and form potential subdimensional motif candidates (stored in $Motifs$). Collaborative Enumerator has four major steps: a SAX word matching step (Line 3-9); a greedy enumeration step (Line 10-12); a dimension matching step (Line 13-15); and a step that updates motif set (Line 16-17). After these four steps, the enumerator recursively calls itself to enumerate longer motifs (Line 18-21).

\begin{algorithm}[t]
    \caption{Collaborative Enumerator (CollabEnum)}
  \begin{algorithmic}[1]

    \STATE \textbf{Input}: Node: $S_{k,j}$, Motif set: $Motifs$

    \STATE \textbf{Output}: Updated Motif Set $Motifs$
    
    {\color{blue}SAX Word Matching Step}
    
    \STATE $S_{observed}$=Merge($S_{k,j}$,$S_{k,j}^{(next)}$)
    \STATE $word$=1D-SAX($S_{observed}$)

    \IF{SAXTable.NotExist($word$,$S_{observed}$,$j$)}
        \STATE SAXTable.put($word$,$j$,$S_{observed}$.length,$S_{observed}$)
        \STATE \textbf{return} $Motifs$
    \ENDIF

    \STATE $S_{match}$=SAXTable.getSimLen($word$,$j$);
    
    {\color{blue}Greedy Enumeration Step}

    \STATE $S_{observed}',S_{match}'=$Enumeration($S_{observed}$,$S_{match}$)  
    \STATE InsertNode($S_{observed}'$);
    
    \STATE InsertNode($S_{match}'$)
    
        {\color{blue}Dimension Matching Step}

    \STATE $\mathbf{d},\mathcal{S}^{o,m}$=MatchDimension($S_{observed}'$,$S_{match}'$) 
    
    \STATE LabelFirstNode($\mathbf{d}$,$S_{observed}'$)

    \STATE $\mathbf{S}_{\mathbf{d}}^{(o)},\mathbf{S}_{\mathbf{d}}^{(m)}$=GenMatchedSeq($\mathbf{d}$,$S_{observed}'$,$S_{match}'$)
    \STATE $\mathbf{wordSet}$=$\mathbf{S}_{\mathbf{d}}^{(m)}$
    
    \STATE $Motifs$.put($\mathbf{wordSet}$,$\mathbf{S}_{\mathbf{d}}^{(o)}$,$\mathbf{S}_{\mathbf{d}}^{(m)}$)

  {\color{blue} Recursively Enumerate Long Subsequence}
  
    \STATE $Motifs=$CollabEnum($S_{observed}'$,$Motifs$);

    \IF{!$S_{match}'$.isEnum()}
    \STATE $Motifs=$CollabEnum($S_{match}'$,$Motifs$);
    \ENDIF
    \STATE return $Motifs$
  \end{algorithmic}
\end{algorithm}

\subsubsection{SAX word Matching}

In this step, Collaborative Enumerator attempts to detect a pair of matching subsequences based on SAX word representation. Specifically, given a node $S_{k,j} \in \mathcal{S}_{NR,j}$, we first merge the two subsequences stored in $S_{k,j}$ and $S_{k,j}^{(next)}$ respectively and compute a SAX word $word$ via the FastSAX algorithm introduced in Sec. IV.A.1. A new node $S_{observed}$ is generated to represent the new merged subsequence (Line 3-4). The SAX word $word$ along with the length of the merged subsequence and its dimension $j$ are inserted into a SAX word table, SAXTable, if the same SAX word representing some subsequence(s) of similar length at dimension $j$ does not already exist in SAXTable (Line 5-6). Otherwise it indicates that the enumerator has found a pair of matching subsequences in dimension $j$, in which case the algorithm gets the node representing the matched subsequence, $S_{match}$, from the SAXTable (Line 9). 

Intuitively, this step looks to see whether any subsequence stored in SAXTable is similar to the newly formed long subsequence. If it finds one successfully, then the algorithm calls local enumeration and dimension matching steps to detect motifs (see below). Otherwise it puts the subsequence into SAXTable for future matching.

\subsubsection{Local Enumeration}

If the algorithm finds matching subsequences in the previous step, CHIME then conducts a local greedy enumeration step to expand the subsequences simultaneously as much as possible to find the longest matching subsequences pair. This pair of subsequences is obtained by continuing merging nodes via the next edge. The process stops when the two subsequences are represented by different SAX words (Line 10). Two nodes, $S_{observed}'$ and $S_{match}'$, are generated to represent the expanded subsequences. The nodes are inserted into $\mathcal{S}_{NR,k}$ for future enumeration (Line 11-12). 

\begin{figure}[h]
\vspace{-2mm}
 \centering
 \includegraphics[width=70mm]{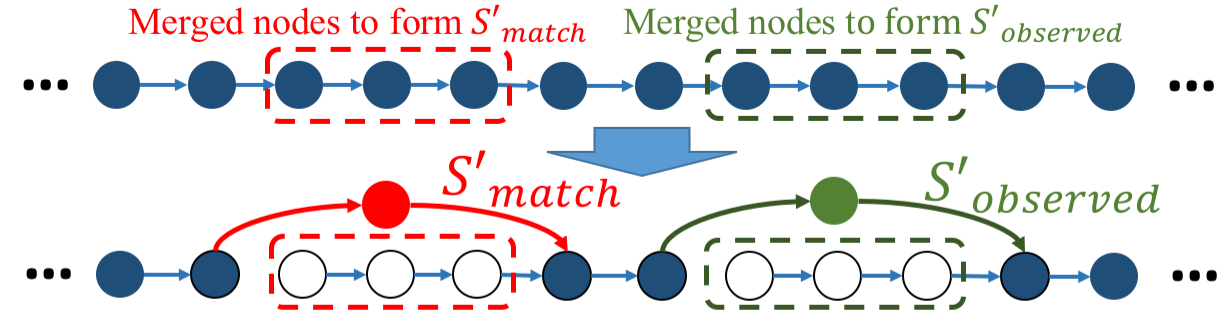}

 \caption{Illustration of Updating Edges}
\vspace{-4mm}
 \label{fig:link}
\end{figure}

Upon insertion of the nodes  $S_{observed}'$ and $S_{match}'$, the edges are updated accordingly as shown in Fig. \ref{fig:link}. Intuitively, the newly inserted nodes allow us to re-use the detected matching subsequences to reduce the cost of generating long subsequences.



\begin{figure}[h]
\vspace{-2mm}
 \centering
 \includegraphics[width=80mm]{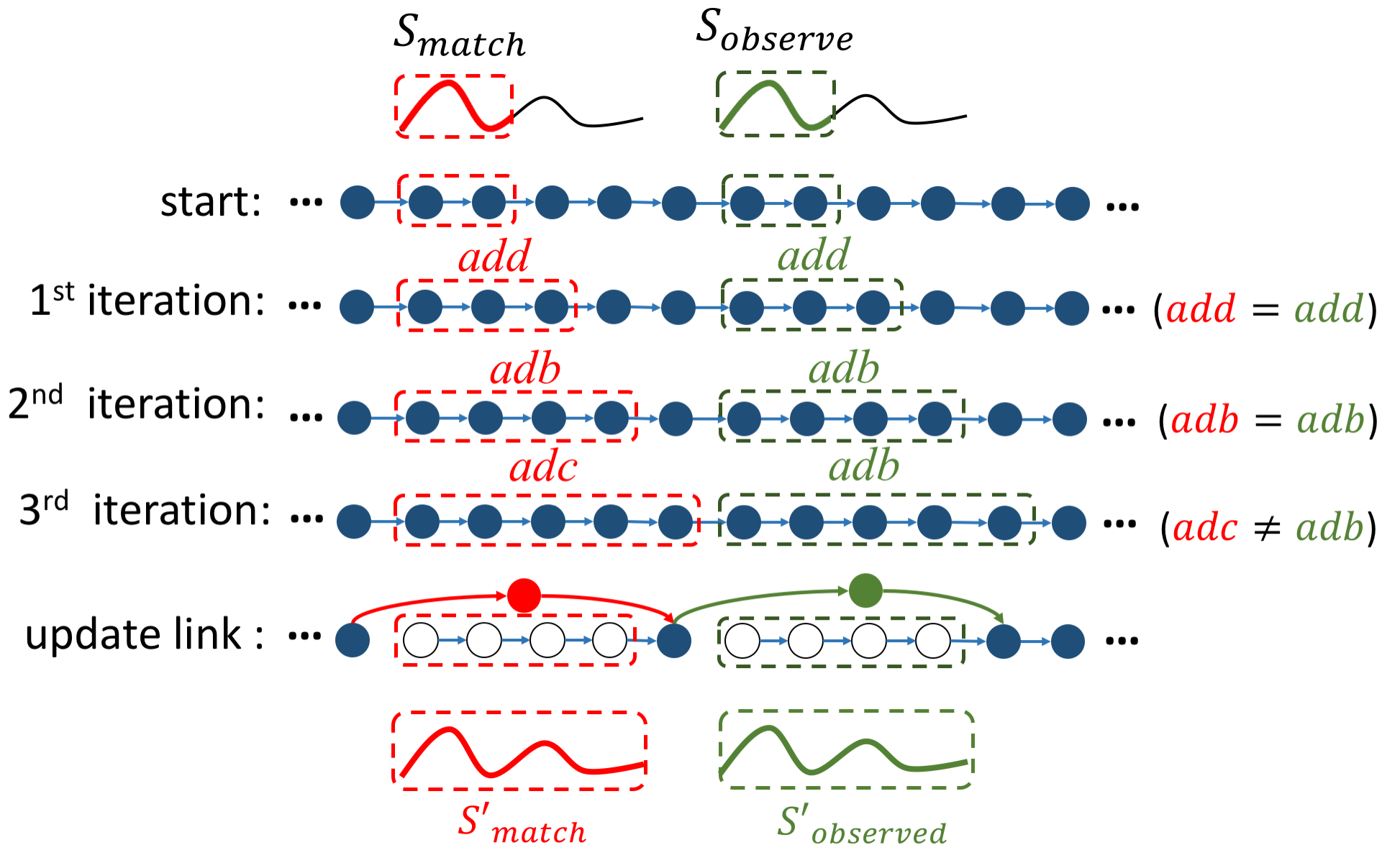}

 \caption{Example of Local Enumeration Step}
 \label{fig:enum}
\end{figure}

An example is shown in Fig. \ref{fig:enum}. The algorithm iteratively merges nodes via the \textit{next} edge in the first and second iterations since in both iterations, the SAX words that represent the two newly generated subsequences are identical (\textit{add} for the first iteration, and \textit{adb} for the second iteration). In the third iteration, the SAX words for the two subsequences are different (\textit{adc} and \textit{adb}, respectively), so the algorithm stops the enumeration. $S_{observed}'$ and $S_{match}'$ (green and red nodes) that represented the green and red long subsequences are formed based on the merged nodes. 

In the local enumeration step, all merged short subsequences are completely overlapped with the longest matched subsequences. As demonstrated in \cite{mueen2013enumeration}\cite{hime}, these covered subsequences are redundant. So CHIME skips these short subsequences without generating nodes to reduce memory cost.

\subsubsection{Dimension Matching}

After the local enumeration step, the algorithm then conducts a dimension matching step. In this step, for each dimension $1\leq d \leq D, d\neq j$, the enumerator compares the corresponding pair of SAX words generated from the subsequences at the same location, and with the same length as $S_{observed}'$ and $S_{match}'$, stored in $\mathcal{S}_{NR}$. An indicator vector $\mathbf{d}$ stores the dimensions in which matching subsequences are found, and a set of nodes $\mathcal{S}^{o,m}$ storing all newly matched subsequences are generated (Line 13). The first nodes of the matching subsequences are marked as ``visited'' to avoid revisiting them in the future. 



\begin{figure}[h]

 \centering
 \includegraphics[width=75mm]{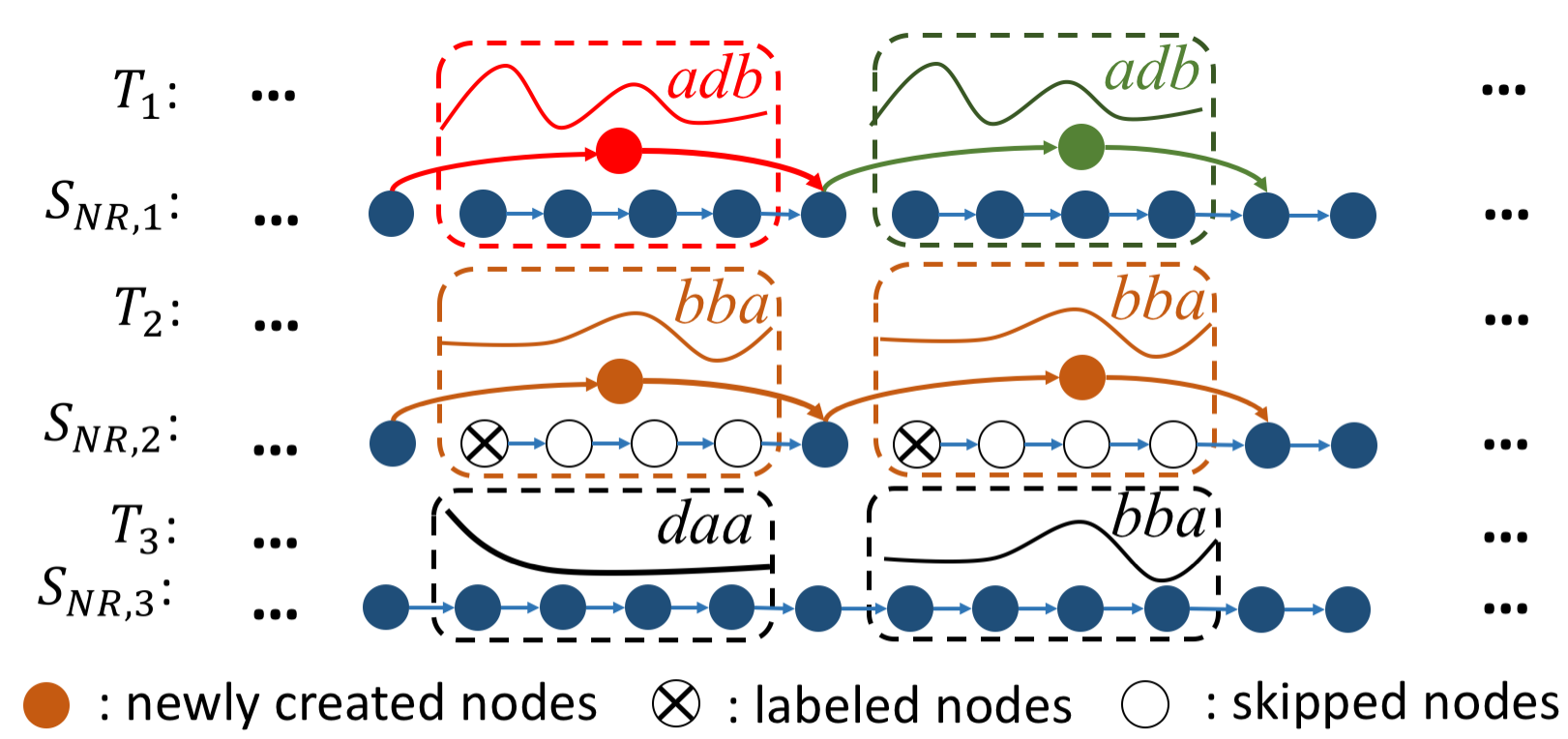}
 \caption{Example of Dimension Search Step.}
 \label{fig:dm}
\end{figure}

To clarify how dimension search process works, let us consider the example shown in Fig. \ref{fig:dm}. Suppose $S_{observed}'$ and $S_{match}'$ represent a pair of matching subsequences in $T_1$ of a three-dimensional time series. In the dimension matching step, CHIME checks each of the remaining two dimensions $T_2$ and $T_3$ and see if the pair of subsequences at the same locations as $S_{observed}'$ and $S_{match}'$, respectively, also have matching SAX words. In this example, CHIME finds that the SAX words match in $T_2$ (subsequences share the same word $bba$). So an indicator vector $\mathbf{d}=\{1,2\}$ and the nodes representing the newly found matching subsequences pair (brown nodes) are formed. The node representing the first covered subsequence (blank nodes with `x') is marked as visited. These two brown nodes are stored in SAXTable for future enumeration.   

Through the dimension matching process, CHIME can directly find matching long subsequences without going through the process of merging all the blank nodes shown in Fig. \ref{fig:dm}, which can reduce the cost. In this example, a trivial solution would need to repeat SAX word matching 4 times to find the long subsequences in $T_2$, whereas CHIME only does it 2 times.

\subsubsection{Update Motif Set}

In this step, CHIME first forms a pair of matching subdimensional subsequences, $\mathbf{S}_{\mathbf{d}}^{(o)}$ and $\mathbf{S}_{\mathbf{d}}^{(m)}$, based on $\mathbf{d}$ and all matching subsequences (Line 16). $\mathbf{S}_{\mathbf{d}}^{(o)}$ and $\mathbf{S}_{\mathbf{d}}^{(m)}$ can be considered as two instances of a subdimensional motif since the SAX words in dimension $j \in \mathbf{d}$ are matched. So CHIME computes a discrete representation by concatenating all the SAX words along with the dimension index to represent these two subdimensional subsequences (e.g. consider the example in Fig. 6, a SAX word sequence $adb1-bba2$ is generated). $\mathbf{S}_{\mathbf{d}}^{(o)}$ and $\mathbf{S}_{\mathbf{d}}^{(m)}$ are then put into the candidate motif set $Motifs$ along with the hash value generated by $\textbf{wordSet}$ (Line 18) and its length. Intuitively, since all subdimensional subsequences represented by the same $\textbf{wordSet}$, with similar lengths are considered instances of the same motif, these subdimensional subsequences will be put into the same bucket in $Motifs$ to form a motif candidate. In the post-processing step, a pairwise comparison is conducted to filter out any false positive candidates.



\subsubsection{Recursive Enumeration}

Finally, CHIME recursively calls itself to enumerate longer motifs. More specifically, the algorithm calls itself to test two newly generated nodes $S_{observed}'$ and $S_{matched}'$ to continue enumerating motifs. Note that since there is a chance that the matched subsequence may already be enumerated into long subsequence in the previous step, $S_{matched}'$ only conducts the recursive enumeration if it has not been done before. The algorithm stops when there is no SAX word matching detected (Line 19-21).

\subsubsection{High-level Framework}

\begin{algorithm}[h]

    \caption{High-level Framework}
  \begin{algorithmic}[1]
  \STATE Input: $\mathcal{S}_{NR}$, Output: $Motifs$
    \STATE $Motifs=\{\}$; SAXTable=$\{\}$;

    \FOR{all $S_{k,j} \in \mathcal{S}_{NR,j}$ from left to right, $j$ from 1 to $D$}
    
        \IF{!$S_{k,j}$.isVisited()} 
        
        \STATE $Motifs$=CollabEnum($S_{k,j}$,$Motifs$)
        \ENDIF
    \ENDFOR
    
    \STATE $Motifs$=RemoveFalsePositive($Motifs$);
  \end{algorithmic}
\end{algorithm}

The overall framework is shown in Alg. 3. Intuitively, CHIME conducts a left to right scan through every  $\mathcal{S}_{NR,j} \in \mathcal{S}_{NR}$ and calls the collaborative enumerator to detect motifs. During the scan, CHIME utilizes the recorded nodes to avoid repeating enumeration of the observed node(s). Specifically, for each node $S_{k,j}$, CHIME only calls enumerator with node $S_{k,j}$ to start the enumeration process if $S_{k,j}$ is not visited (Line 5). If $S_{k,j}$ is previously visited, it indicates that the corresponding subsequence has already been processed, so CHIME skips the node to avoid redundant work. Finally, after all nodes are enumerated, a post-processing is conducted to remove all false positive instances of motifs stored in $Motifs$ (Line 8). In the proposed work, since we compare the scalability of CHIME with state-of-the-art approach \cite{yeh2017matrix}, we compute the distances between pairs of instances that share the same $\mathbf{wordSet}$ representation and have similar lengths. We then rank the motifs by distances in ascending order per dimension size per motif length. Note that the cost is the same as filtering out all false positive instances detected by CHIME per the definition of subdimensional motif in this work. 
\subsection{Time Complexity}

The time complexity of CHIME is dominated by pairwise comparison post-processing, which may take $O(DN^2L_{max})$ time, similar to state-of-the-art motif enumeration approach. However, since CHIME utilizes symbolic representation to avoid enumerating many subsequences redundantly, in the experiments, we show that the algorithm can indeed detect subdimensional motifs with very small amount of time. It can also handle million size multivariate time series, which none of the existing approaches can handle due to the time cost.

\subsection{Compared with Sequence Matching Approach}

One existing work \cite{balasubramanian2016discovering} utilizes sequence matching approach (e.g. \cite{balasubramanian2013flexible}\cite{nevill1997identifying}) to detect variable-length subdimensional motif. While the approach cannot handle high dimensional multivariate time series due to the exponential time and space requirements toward dimension size, it is worth noting that even if the algorithm could reduce the time cost to a reasonable cost, the algorithm still cannot fulfill the task introduced in this paper.

\begin{figure}[h]
\vspace{-2mm}
 \centering
 \includegraphics[width=75mm]{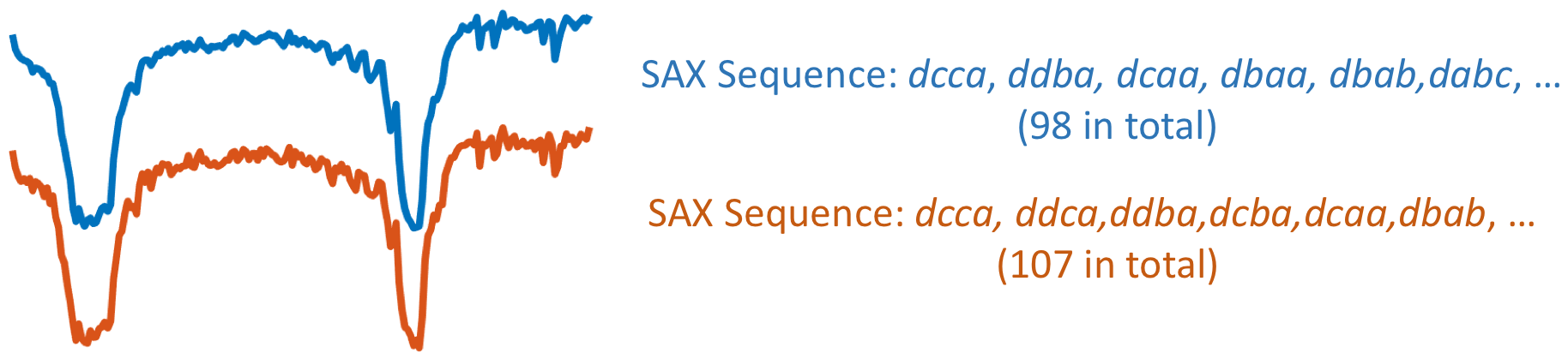}
 \caption{Example of Two Similar Sequence and Correspond SAX Sequence given $w=4,a=4$}
 \label{fig:sq}
\end{figure}

The problem is twofold. First, long subsequences are typically represented by overwhelmingly long sequences of symbols. For example, consider the two 1-D subsequences of length 200 \cite{pems}, shown in Figure \ref{fig:sq}. Visually, these two time series look very similar. The sequence representations of the time series obtained by using sliding window of length 20 (10\% of the subsequence length) is shown on the right. We can see that the two SAX sequences, including their lengths, look very different despite the striking similarity between the time series. Since the previous approach \cite{balasubramanian2016discovering} is based on SAX sequence matching \cite{nevill1997identifying}, it would not be able to find the motifs unless a warping robust matching process is conducted, which is too costly given the complexity. In contrast, CHIME recomputes the SAX words and can represents these two 1-D sequences by one SAX word, which is robust to noise.

Second, the mean and variance of a short subsequence used to generate the SAX word sequence can significantly differ from that of a long subsequence \cite{mueen2013enumeration}\cite{valmod}. Since the shape can be largely affected by mean and variance, the SAX word sequence based representation may fail to capture the actual similarity between two subsequences. As a result, SAX word sequence matching based approaches such as \cite{itr}\cite{li2012visualizing}\cite{Gao2018} only can detect motifs in a small length range, whereas CHIME does not have this problem.   


\section{Experiments}

We perform a series of experiments to evaluate the accuracy and speed of CHIME. All the experiments are conducted on a 16 GB RAM laptop with quad core processor of 2.5 GHz. The executable software and datasets used in the experiments can be found in \href{https://github.com/flash121123/CHIME}{\textit{https://github.com/flash121123/CHIME}}. In all evaluation experiments unless noted, the PAA parameters $w$ and $a$ are set to 5 and 6, respectively. The minimum length $l$ is 300 and motif threshold is $R(L)=0.02L$.  



We first demonstrate that existing work may not be suitable for variable-length subdimensional motif discovery. As shown in previous work \cite{mueen2013enumeration}\cite{valmod}\cite{Gao2018}, index-based fixed-length \textit{approximate} motifs discovery algorithms such as random projection \cite{chiu2003probabilistic} are not suitable for detecting variable-length motifs due to memory requirement. This is because the algorithm needs to generate discrete representation for every subsequence for every length tested, and it can soon become impractical even with small enumeration range\cite{mueen2013enumeration}\cite{Gao2018}. To demonstrate the significant difference in memory requirement between the two algorithms, we conduct a simple experiment. We compare the ratio between the number of subsequences stored in memory for matching motifs, and the product of the number of subsequences and enumeration range:

\begin{equation}
    ratio=\frac{\# of Subsequences Kept Track}{\# of Subsequences \times Enumeration range}
\end{equation}

Since the random projection approach \cite{minnen2007detecting} discretizes and keeps track all tested subsequences of every length, the ratio is equal to 1.

The ratios in three different types of datasets --- random walk data, traffic speed data\cite{pems}, and EEG data\footnote{http://bbci.de/competition/iv/} along with data size and the enumeration range (measured by number of distinct motif length detected after removing all false positive) are shown in Table 1. In all three datasets, the memory cost for CHIME is three orders of magnitude lower compared with random projection. This property allows CHIME to detect variable-length subdimensional motifs in large scale datasets.

\begin{table}[h]
\vspace{-2mm}
\caption{CHIME Vs. Random Projection in Memory Cost}
\vspace{-2mm}
\centering
\scalebox{0.8}{
\begin{tabular}{ |c|c|c|c|c| }
  \hline
  Dataset  & Size ($D\times N$) & Enumeration Range & CHIME & Random Projection \\
    \hline
  Random Walk &  $50\times 10^6$ & 8291 & \textbf{0.001} & 1  \\
   \hline
   Traffic Speed & $10\times 10^5$ & 4177 &  \textbf{0.0014} & 1\\
    \hline
   EEG  & $10\times 10^5$ & 1674 & \textbf{0.004} & 1 \\
   \hline
\end{tabular}
}
\vspace{-2mm}
\label{tab:memory}
\end{table}


Since there is no comparable approximate variable-length subdimensional motif discovery approach that can solve the problem in the tested scale, we compare with the state-of-the-art variable-length motif discovery solution introduced by Nunthanid et al. \cite{nunthanid2011discovery} --- the algorithm conducts a brute-force enumeration process to enumerate motifs of different lengths via a fixed-length motif discovery approach. We demonstrate that the scale of problems handled by CHIME is too large to get exact solution. 


\subsection{Detecting Planted Motifs in Random Walk Time Series}

\begin{figure}[h]    
\centering
    \begin{subfigure}[b]{40mm}
        \includegraphics[width=\textwidth]{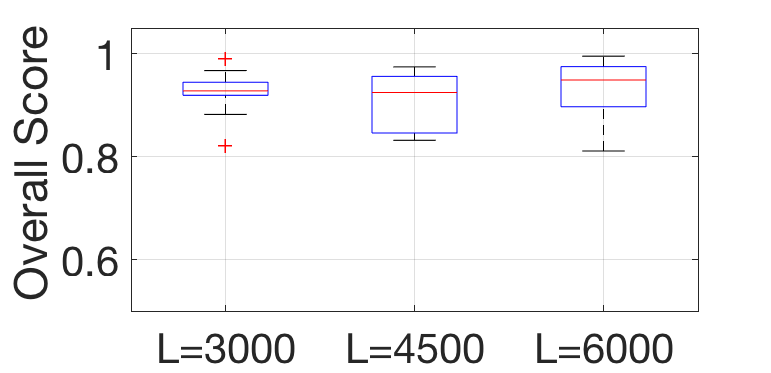}

        \caption{\footnotesize Overall Score Vs. $L$}
        \label{fig:a}
    \end{subfigure}
    \begin{subfigure}[b]{40mm}
        \includegraphics[width=\textwidth]{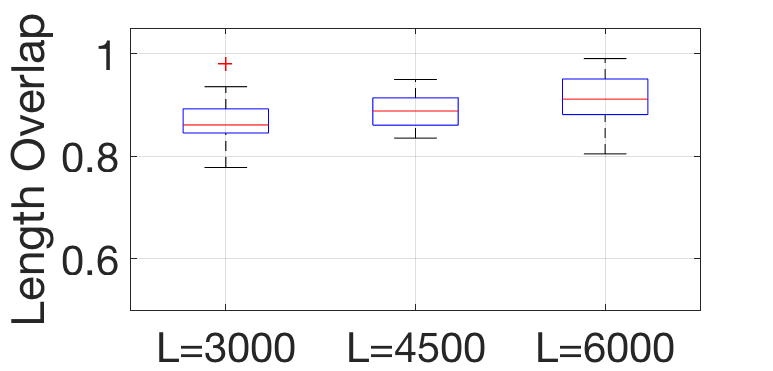}
 
        \caption{\footnotesize Length Overlap Vs. $L$}
        \label{fig:b}
    \end{subfigure}
    \begin{subfigure}[b]{40mm}
        \includegraphics[width=\textwidth]{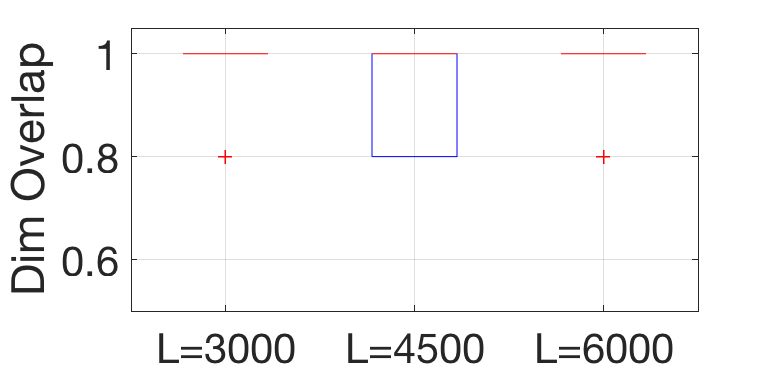}

        \caption{\footnotesize Dimension Overlap Vs. $L$}
        \label{fig:c}
    \end{subfigure}
    \begin{subfigure}[b]{40mm}
        \includegraphics[width=\textwidth]{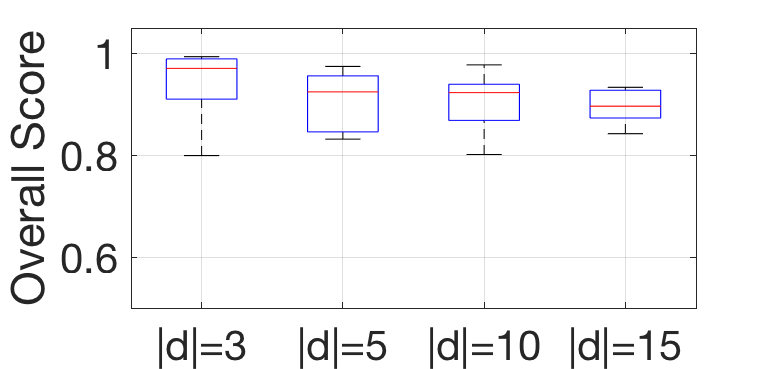}
       
        \caption{\footnotesize  Overall Score vs $|\mathbf{d}|$}
        \label{fig:a}
    \end{subfigure}
    \begin{subfigure}[b]{40mm}
        \includegraphics[width=\textwidth]{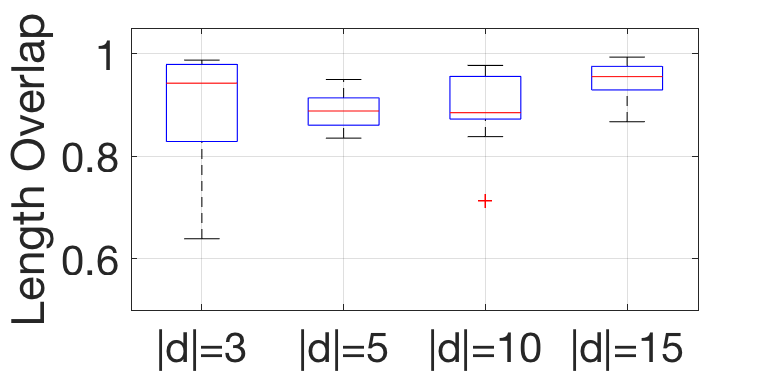}
    
        \caption{\footnotesize  Length Overlap vs. $|\mathbf{d}|$}
        \label{fig:b}
    \end{subfigure}
    \begin{subfigure}[b]{40mm}
        \includegraphics[width=\textwidth]{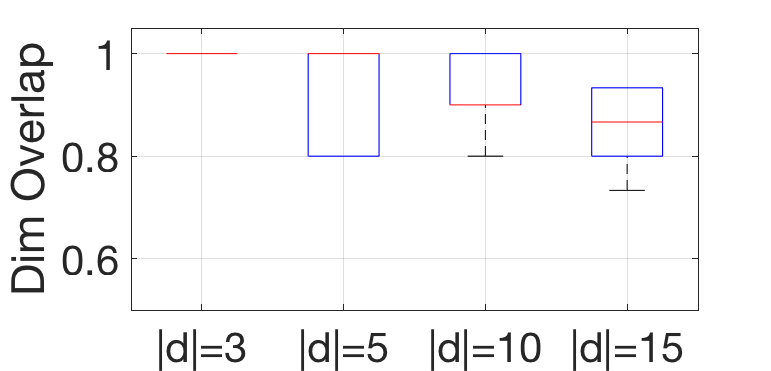}
  
        \caption{\footnotesize  Dimension Overlap vs. $|\mathbf{d}|$}
        \label{fig:c}
    \end{subfigure}
 
\caption{Planted Motif VS. Overlap Score}
\label{fig:plant2}

\end{figure}

We first tested CHIME in a planted motif experiment \cite{Gao2018}\cite{hime}\cite{mueen2013enumeration} to demonstrate its ability to detect subdimensional motifs with high accuracy when the minimum length is much shorter than the actual motif length. 

We planted a subdimensional motif of 10 instances into a random walk time series of length 1 million points, twenty dimensions with random positions and random dimensions. The shape of motifs are generated by using $x=\sum_{i=1}^{5}A_i\sin{\alpha_i x+\beta_i}$ with random parameters $A_i \in [0,5]$, $\alpha_i \in [-2,2]$ and $\beta_i \in [-\pi,\pi]$. We added 5\% random noise to every instance of the motifs. The mean and variance are also randomly generated. CHIME is expected to find at least a pair of non-overlapping subsequences that highly overlap with the actual planted instances. Similar to previous work\cite{hime}\cite{Gao2018}, we evaluate the performance by the overlapping rate with the actual planted intervals and relevant dimensions. We also evaluate the performance via an overall overlap score computed by the geometric average of both metrics. The high overall overlap score indicates that the algorithm found a motif highly overlapped with the planted motif in most of relevant dimensions. 






\subsubsection{Planted Motifs of Different Lengths}

We first tested CHIME with planted motifs of length $L$ equal to 3000, 4500, and 6000. The number of relevant dimensions for the planted motif is set to be 5. We repeat the experiment 10 times for each motif length. The boxplot of overall overlapping rate for each motif length is shown in Fig. \ref{fig:plant2}(a). CHIME consistently gets overall overlapping rates above 0.8 for all motif lengths 3000, 4500 and 6000. CHIME also consistently gets high length overlap rate in all tests. Besides, according to Fig. \ref{fig:plant2}(b)-(c), the algorithm also maintains over 0.8 overlapping rate in both length and dimension.


\subsubsection{Planted Motifs of Different Number of Relevant Dimensions}

In this experiment, we tested CHIME with planted motifs for which the number of relevant dimensions $|\mathbf{d}|$ equals to 3, 5, 10, 15. The motif length is set to be 4500. Similar to the previous experiment, we repeated the experiment 10 times for each dimension setting, and the result is shown in Fig. \ref{fig:plant2}(d)-(f). We observe that the median overall overlapping score maintains at approximately 0.9. While the length and dimension overlapping rate decrease as the number of relevant dimension increased, the median of both metrics are still over 0.8. The result indicates that even if motifs contain a large number of relevant dimension, CHIME  still can find most of relevant dimensions.


\subsection{Scalability}

In this subsection, we conduct experiments to evaluate the scalability over length and dimension of the multivariate time series. Since there is no approximate variable-length motif discovery algorithm that can handle the size of data tested in the experiment, we report the execution time of state-of-the-art fixed-length multidimensional motif discovery algorithm\cite{yeh2017matrix} performance with STOMP\cite{zhumatrix} as the base approach. The code is provided by the authors and written in C. In the test case where the algorithm takes more than 24 hour to complete, we estimate the execution time by the first 1000 iterations (the estimation approach used in previous work \cite{zhumatrix}). We also report the estimated brute-force motif enumeration time if utilizing the framework described in \cite{nunthanid2011discovery}, the classical approach used in variable-length motif discovery. The estimated time is computed by the fixed-length motif discovery time multiplied by the enumeration range since STOMP's execution time is invariant to motif length\cite{zhumatrix}.


\begin{table}[t]
\centering
\caption{Execution Time Vs. Time Series Length in a 50 dimensional Time Series}

\scalebox{0.75}{
\centering
\begin{tabular}{ |c|c|c|c|c|c| }
  \hline
   \multicolumn{1}{|c|}{\centering Time Series Length}   & 200K & 400K & 600K & 800K & 1 million\\
   \hline
   \multicolumn{1}{|c|}{\centering Enumeration Range} & 2724 & 4499 & 5915 & 7319 & 8290\\
   \multicolumn{1}{|c|}{\centering Dimension Range}& 5 & 6 & 6 & 6 & 6\\
   \hline
    CHIME   & \textbf{3.93 min} &  \textbf{10.03 min} & \textbf{15.1 min} & \textbf{22.5 min} &  \textbf{28.05 min}\\
    \multicolumn{1}{|c|}{\centering Post-processing} &  \textbf{54 sec.} &  \textbf{3.4 min} & \textbf{11 min} &  \textbf{19.5 min} &  \textbf{31 min}\\
   \hline
   \hline
   \multicolumn{1}{|c|}{\centering Fixed-length Motif Discovery}  & 3.6 hr & 11.6 hr & \textcolor{gray}{1.04 days} & \textcolor{gray}{2.08 days} & \textcolor{gray}{12 days}\\
   \multicolumn{1}{|c|}{\centering Estimated Brute Force Time}  & \textcolor{gray}{1.19 yr.} & \textcolor{gray}{5.95 yr.} & \textcolor{gray}{16.8 yr.} & \textcolor{gray}{41.68 yr} & \textcolor{gray}{272 yr}\\ 
   \hline
\end{tabular}
}
\label{tab:memory}
\vspace{-3mm}
\end{table}

\subsubsection{Scalability over Time Series Length}

We tested the scalability of CHIME in a one million length random walk time series of fifty dimensions. The growths of execution time, enumeration length range and dimension range as the length increases are shown in Table 2. The enumeration length and dimension range are measured by the number of distinct motif lengths and dimensions detected after removing all false positive instances based on the motif threshold function. According to the table, the execution time for CHIME grows much slower than that of the tested state-of-the-art approaches. In the largest cases, the algorithm takes 28 min to complete, and then another 31 min to do pairwise distance comparisons. The estimated execution time for fixed-length motif discovery approach to detect motifs is 12 days. Similarly, the enumeration length range and dimension range grow as the length of time series grows. In the largest case, the enumeration range has almost 9000 different lengths. Consider the estimated execution time for the brute force approach, the length range enumerated by CHIME is too large for the state-of-the-art algorithm to get exact solution. In contrast, CHIME provides an alternative way to efficiently detect approximate variable-length motifs in this scale of data size.

\begin{table}[h]
\caption{Execution Time Vs. Dimension Size}
\centering
\scalebox{0.77}{
\begin{tabular}{ |c|c|c|c|c| }
  \hline
  Dimension  & 25 & 50 & 100 & 200 \\
  \hline\hline
  CHIME (Total Time) & \textbf{1.1 min} & \textbf{3.4 min} & \textbf{10.5 min} & \textbf{33.5 min} \\
  
  Enumeration Range & 2058 & 2252 & 2334 & 2419\\
  
  \hline\hline
   Fixed-length Motif Discovery Approach  & 1.72 hr & 3.45 hr & 6.9 hr & 13.8 hr \\
    Estimated Brute Force Time & \textcolor{gray}{58.9 days} &  \textcolor{gray}{129.49 days} &  \textcolor{gray}{268 days}&  \textcolor{gray}{1.52 yr}\\
    \hline
\end{tabular}
}
\label{tab:a}
\end{table}

\subsubsection{Scalability Over Number of Dimensions}

We then tested the scalability of CHIME in a 200-dimensional, 160,000 length random walk time series. The growth of execution time as dimension increases is shown in Table 3. The execution time for CHIME grows faster than that of fixed-length motif discovery approach. However, the estimated brute-force execution time is still infeasible due to the large enumeration range. In contrast, CHIME only takes less than one hour to complete. 

\subsection{Parameters Analysis}

\begin{figure}[h]
\vspace{-2mm}
\centering
    \begin{subfigure}[b]{40mm}
        \includegraphics[width=\textwidth]{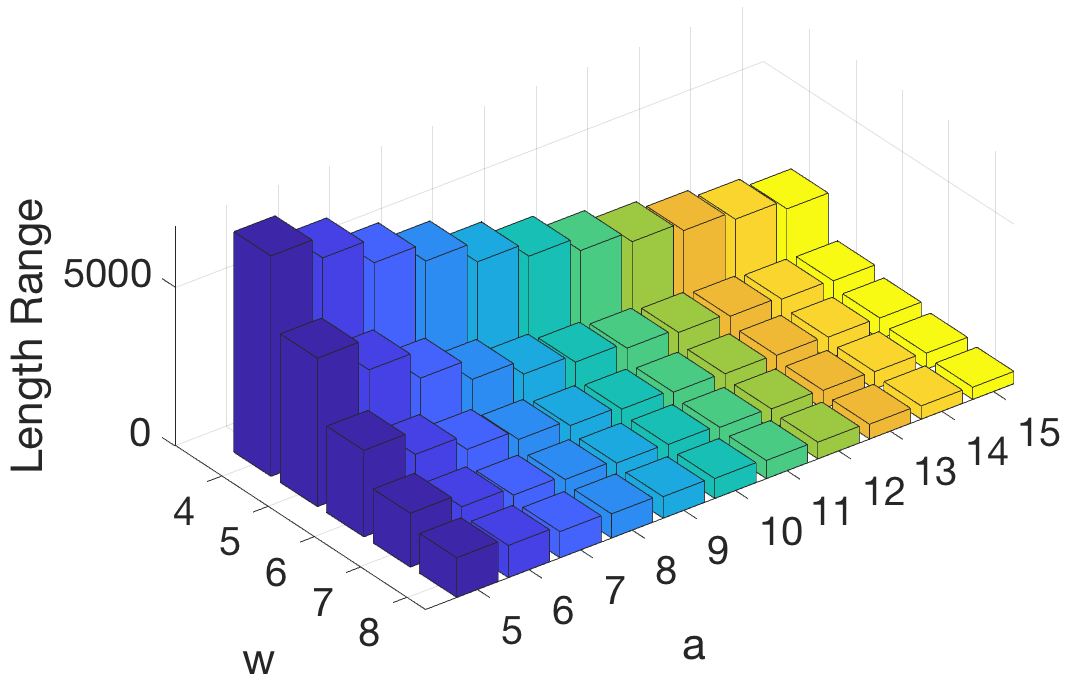}
        \caption{\footnotesize Length Range}
        \label{fig:a}
    \end{subfigure}
    \begin{subfigure}[b]{40mm}
        \includegraphics[width=\textwidth]{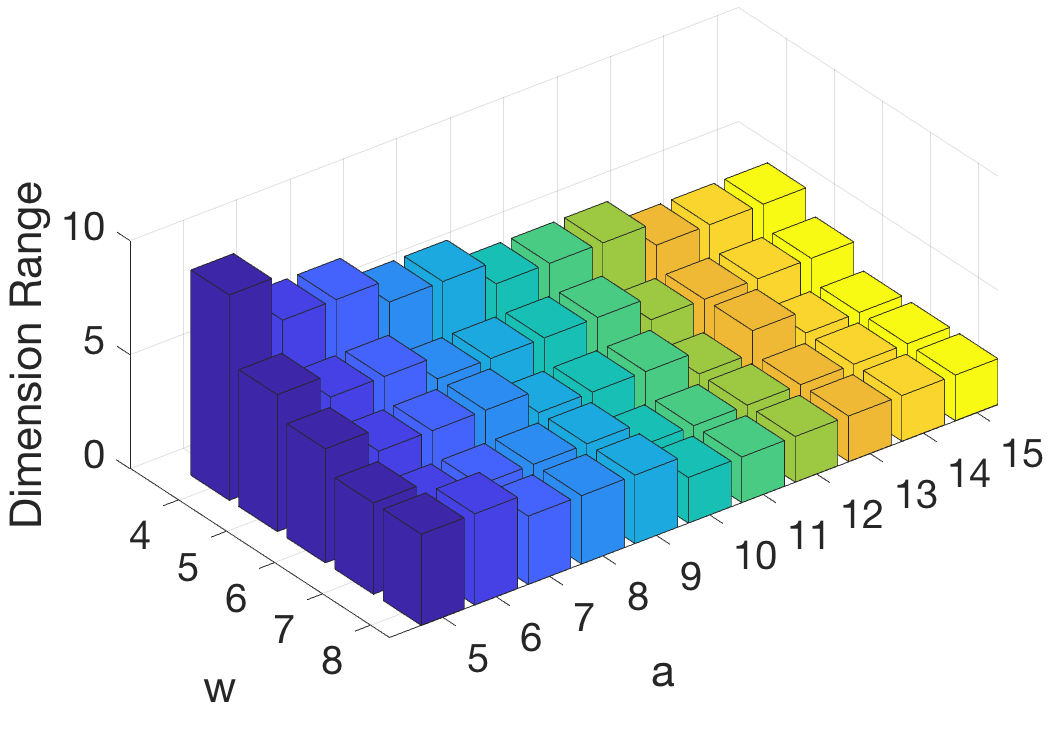}
        \caption{\footnotesize Dimension Range}
        \label{fig:b}
    \end{subfigure}
    \begin{subfigure}[b]{40mm}
        \includegraphics[width=\textwidth]{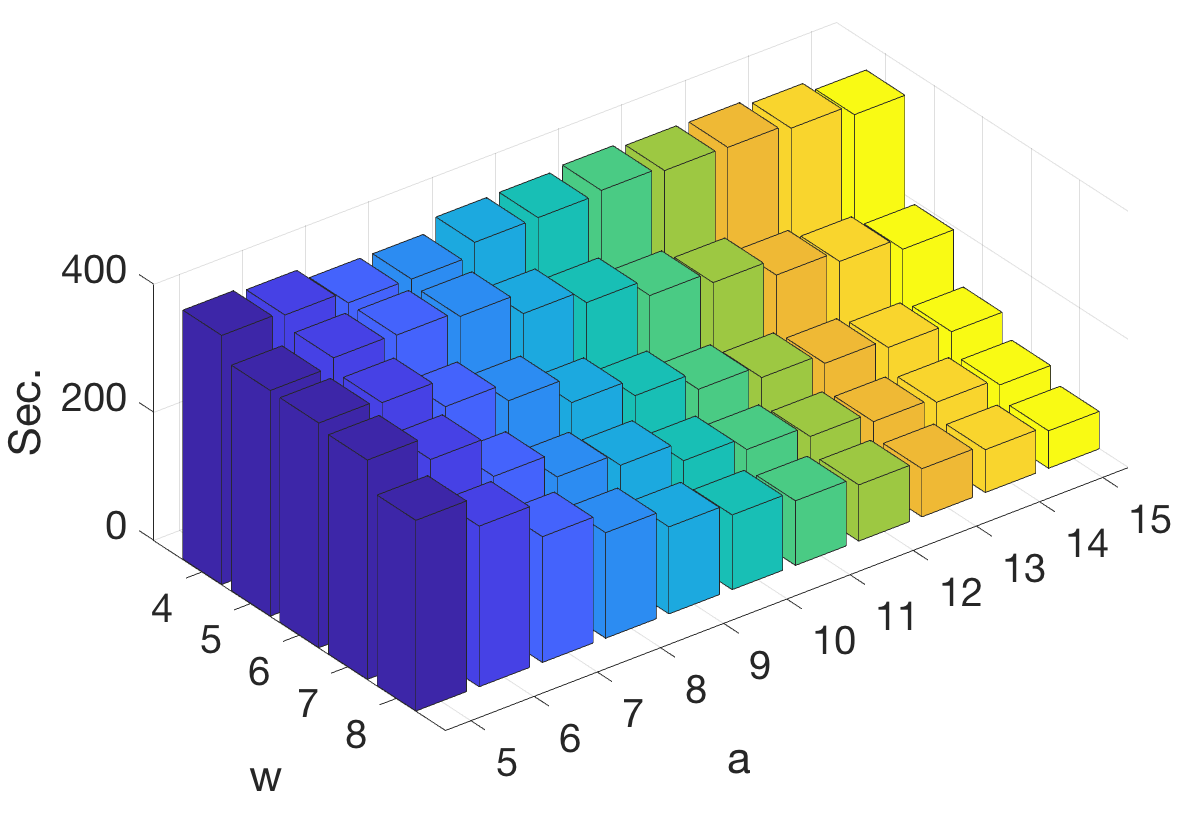}
        \caption{\footnotesize Execution Time}
        \label{fig:c}
    \end{subfigure}

\caption{Parameters Experiments}
\label{fig:param2}
\vspace{-4mm}
\end{figure}

We tested CHIME with parameters $w$ from 4 to 8 and $a$ from 5 to 15 on a 300,000 length, 50-dimensional random walk time series. The enumeration length range, dimension range and execution time of all parameter combinations are shown in Fig. \ref{fig:param2}(a)-(c) respectively. All three values increase as $w$ and $a$ decrease. This is because the number of distinct SAX words increases as $w$ and $a$ increase. As a result, the chance that the SAX words match is reduced. So CHIME can process time series fast at a cost of reduced enumeration length (dimension) range. The user can set both parameters $w$ and $a$ to balance the search ability and execution time.

\subsection{Case Studies}

In this section, we show that CHIME can find high quality motifs in several real world large-scale multivariate time series data.

\begin{figure}[h]    

\centering
    \includegraphics[width=80mm]{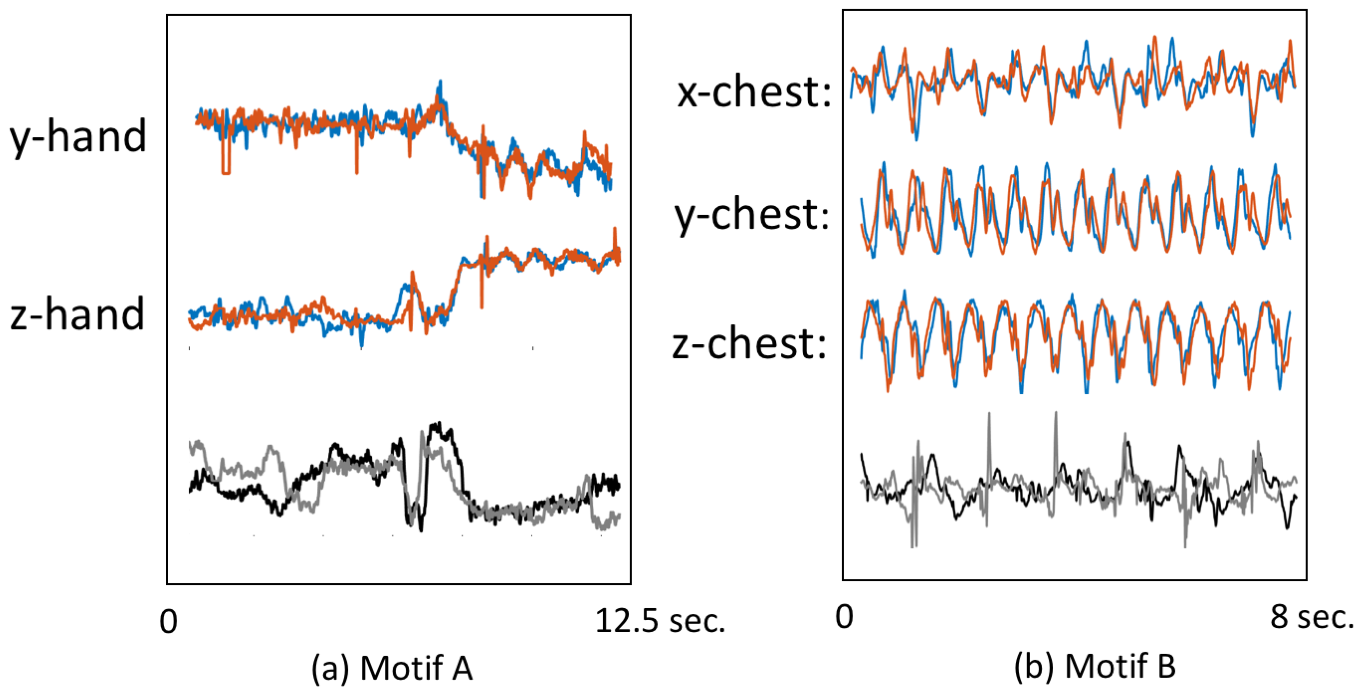}
\vspace{-2mm}
\caption{Two Example of Subdimensional Motifs found in PAMAP2. (Two instances shown in blue and red. One irrelevant dimension are shown in black and grey as an example.)}
\label{fig:pap}
\vspace{-4mm}
\end{figure}

\subsubsection{PAMAP2 Physical Activity Monitoring Time Series}

We first tested CHIME in PAMAP2 Physical Activity Monitoring dataset \cite{reiss2012introducing}. We used all available high resolution acceleration signals recorded from hands, chest and ankles to generate a 9-dimensional time series of 1.3 million points in length. We tested CHIME with minimum motif length equal to 3 sec (300 points). 


Two examples of detected subdimensional motifs are shown in Fig. \ref{fig:pap}(a)(b). The lengths of Motif A and Motif B are 12.5 sec. and 8 sec. respectively. Motif A consists of two different signals (y and z axes of acceleration). Both these signals are recorded from hand motion. According to the label, two occurrences of Motif A indicate an action of ironing, an activity mostly relying on hand. Motif B consists of all 3 signals recorded from chest. According to the annotations, both instances coincide with when the subjects conduct walking activity --- a periodic activity. Both motifs have the ability to explain the patterns that occur in different types of activities. Such information can provide useful insights for behavioral studies. Clearly, since the two motifs have significant length gap (8 sec. vs. 12.5 sec.), repeated execution of fixed length subdimensional motif discovery approach \cite{yeh2017matrix} to enumerate longer motifs is very time consuming given the data size ($9\times1.3 million$).

\subsubsection{Electric Power Demand Time Series}

\begin{figure}[h]

 \centering
 \includegraphics[width=75mm]{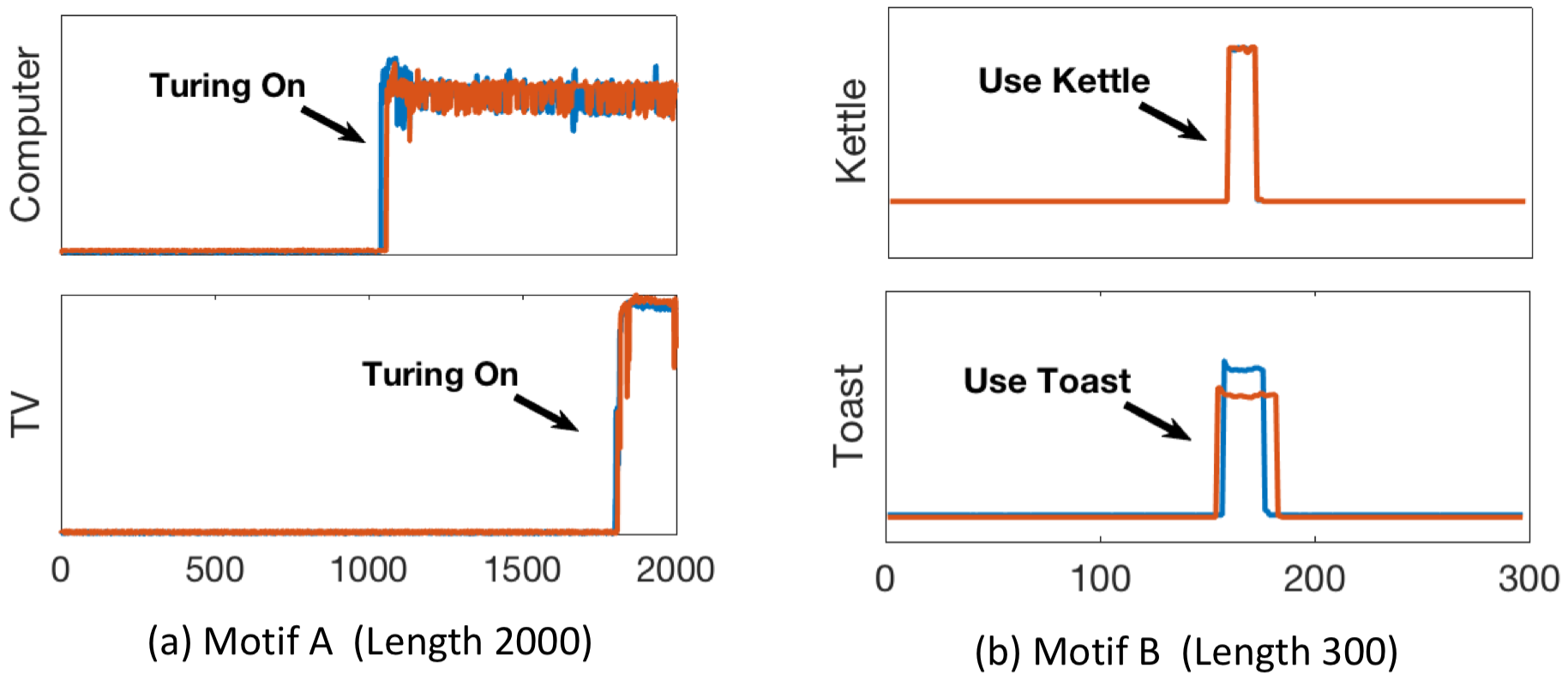}

 \caption{Two Example of Subdimensional Motifs found in Power Usage Time Series (Two instances shown in blue and red).}
 \label{fig:power}

\end{figure}

Interpreting the behavior of Electric power usage has many potential applications\cite{yeh2017matrix}. Recent work shows that motifs can be used to understand activities in this type of data. In this experiment, we apply CHIME on the 1 million length electric power usage dataset \cite{murray2015energy}. The multivariate time series consists of power usage watt of eight different appliances including Washing Machine, Dryer, Dishwasher, Computer Site, Television Site, Combination Microwave, Kettle and Toaster, and is recorded from 2013-Oct. to 2014-Jan. We set the minimum motif length equal to 200 (20 min). Two examples of subdimensional motif are shown in Fig.\ref{fig:power}. 

The first motif of length 2000 is a subdimensonal motif consisting of two appliances --- power usage time series collected from computer site and television site. The motif represents a power usage pattern that the user turns on the computer first and then turns on the TV. CHIME successfully captures this repeating pattern among the data even when the minimum motif enumeration length is a lot smaller than that of the pattern detected. CHIME also discovers the second motif of length 300. This motif consists of two appliances: Kettle and Toaster. Both appliances' power usage patterns are much shorter, and only last several minutes. The two motifs have significantly different motif lengths and represent different, meaningful power usage patterns. Since the time series is large (8 x 1 million in size), repeatedly running a fixed-length motif discovery algorithm such as \cite{yeh2017matrix} will be very time consuming.

\subsubsection{Interpretable Time Series Classification Via Motif}

\begin{figure}[h]

 \centering
 \includegraphics[width=85mm]{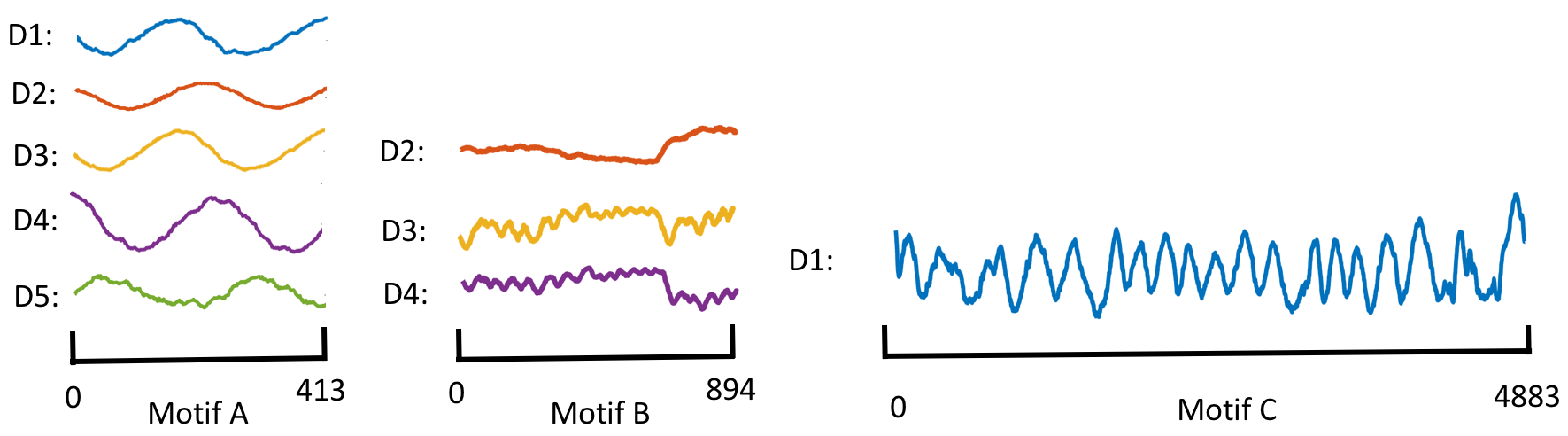}
 \caption{Three Examples of Shapelets Selected via Forward Feature Selection based on CHIME's result} 
 \label{fig:worm}

\end{figure}

One popular application of motif discovery is learning interpretable time series classification model using motifs or shapelets \cite{wangrpm}\cite{gao2018hime}. It has been demonstrated in previous work \cite{wangrpm}\cite{mueen2013enumeration} that motifs that can be used to distinguish between different classes often have different lengths. 

In this experiment, we tested our algorithm in the EigenWorms dataset, a dataset that collects approximately 400 worms' trajectories. Each trajectory consists of approximately 18000 sample points, and each record contains 6 dimensions representing 6 different worm movement features. We ran CHIME on the data with minimum motif length set to 300 sample points. Similar to a previous work on univariate time series classification \cite{wangrpm}, after identifying the motifs, we transformed each original time series to a distance feature vector by computing the closest match distance between the time series and each of the detected subdimensional motif. Then a forward feature selection process is applied to select the features that can achieve the best accuracy via a simple decision tree classifier. We find that by using the three motifs shown in Figure \ref{fig:worm}, the decision tree can already get approximately 70\% accuracy. By further increasing the number of motifs, we find that we can use six motifs to achieve 77\% accuracy in the dataset, whereas 1-Nearest-Neighbor classifier with Dynamic Time Warping only can get 60\% accuracy according to Bagnall et al. \cite{bagnall2018uea}. 


\section{Conclusion}
We introduce a new algorithm, CHIME, to detect approximate subdimensional motifs of different lengths in multivariate time series. CHIME can handle large-size multivariate time series that the state-of-the-art exact algorithms cannot handle efficiently. We show that CHIME can detect subdimensional motifs successfully even when the motif length is considerably larger than the minimum length. In the case studies, we demonstrate that the motifs found by CHIME are meaningful and can potentially have significant impacts in various applications.

\bibliographystyle{IEEEtran}
\bibliography{chime}
\end{document}